# Using Artificial Neural Networks (ANN) to Control Chaos

Dr. Ibrahim Ighneiwa[a] *, Salwa Hamidatou[a], and Fadia Ben Ismael[a]

[a] *Department of Electrical and Electronics Engineering,
Faculty of Engineering, University of Benghazi, Benghazi, Libya*

∗ *218-92-4036057; e-mail: Ibrahim.ighneiwa@uob.edu.ly*

## ABSTRACT

Controlling Chaos could be a big factor in getting great stable amounts of energy out of small amounts of not necessarily stable resources. By definition, Chaos is getting huge changes in the system's output due to unpredictable small changes in initial conditions, and that means we could take advantage of this fact and select the proper control system to manipulate system's initial conditions and inputs in general and get some desirable outputs out of undesirable Chaos. In this work, Artificial Neural Networks (ANN) techniques were used to get a desirable output out of otherwise a Chaotic system. That was accomplished by first building some known chaotic circuit (Chua circuit) and then NI's MultiSim was used to simulate the ANN control system. It was shown that this technique can also be used to stabilize some hard to stabilize electronic systems.

## 1. Introduction

Chaotic systems control have received the interest of many researchers [1], due to the fact that chaotic systems, which are known for its Undesirable behavior, could be beneficial in many areas of science and technology. It has many potential applications [2], such as heat transfer, biological systems, laser physics, chemical reactor, biomedical, economics, weather, and secure communication. Many methods for controlling chaos have been developed, most of which using classic control techniques, such as linear feedback method, active control approach, adaptive technique, nonlinear controller [1,3]. New methods in chaos control were developed by utilizing intelligent control, such as Artificial Neural Networks (ANN), Fuzzy Logic, Genetic Algorithm and Genetic Programming, to name a few [4].

In this work ANN intelligent technique was used to control chaos in electronic circuits. A well known chaos model, namely Chua circuit model, was used to implement such technique. After building the electronic circuit, the ANN control was applied to it and through changing ANN weights of the circuit control it was verified that some desirable outputs could be accomplished.

This paper is organized as follows. Section 2 is about Chaos and its mathematical aspects and how it could be shown both as a real output on an oscilloscope and as a simulated output of some known electronic circuits such as Chua circuit. Section 3 is a discussion of Artificial Neural Networks (ANN), which was used to control chaos. It also discuss the different parts of the network and its various mathematical functions. Section 4 is a discussion of the ANN techniques that was used to control chaos and show that this work leads to getting some desirable stable outputs out of a chaotic system. The final section contains some conclusions and some suggestions that could lead to improving the efficiency of the controlled chaotic system.

## 2. Chaos

Chaos is defined as the property of some non-linear dynamic systems which exhibit sensitive dependence on initial conditions and it is happening when the smallest of changes in a system results in very large differences in system's behavior. [5,14] The so-called butterfly effect has become one of the most popular images of chaos; the idea is that the flapping of a butterfly's wings in Argentina could cause a tornado in Texas [6].

Chaos does not mean that things happen at random, chaos is orderly disorder and not random at all, it is unpredictable in the sense that you cannot predict in what way the system's behavior will change for any change in the input [7].

### 2.1. Lorenz Oscillator

A good example of Chaos is the so called Lorenz Oscillator, where three ordinary differential equations (ODEs) define the chaotic behavior of such Oscillator. These equations are [8]:

$dx/dt = \delta (y - x)$    (1)

$dy/dt = x ( \rho - z ) - y$    (2)

$dz/dt = x y - \beta z$    (3)

Where x,y and z define the state of the system, t is time, and δ, ρ and β are system parameters

Generally, the system does not show any kind of chaotic behavior, But for certain values of its parameters like: β = 8/3, δ = 10, ρ = 28, the system may produce the following chaotic diagram (Fig. 1) [9].

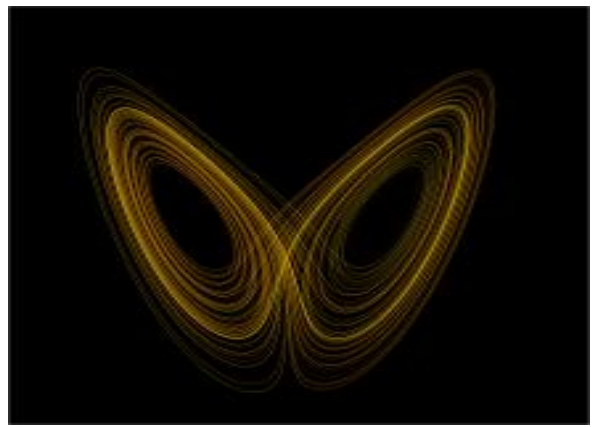

Fig. 1.  A plot of the Lorenz attractor
for values ρ = 28, σ = 10, *b* = 8/3

Another good example of Chaos occurs when increasing µ in the logistic map equation ($x_{n+1} = \mu \, x_n \, (1 + x_n)$) beyond 3.3. [9] For µ<3.3, the system oscillates between two values of x (period-2 cycle). Increasing µ any further, makes the system oscillate between four values (bifurcation/period-4 cycle). And as proved in [9], if we keep increasing µ, period-doublings will happen at smaller intervals of parameter µ (see Fig. 2 below). [10]

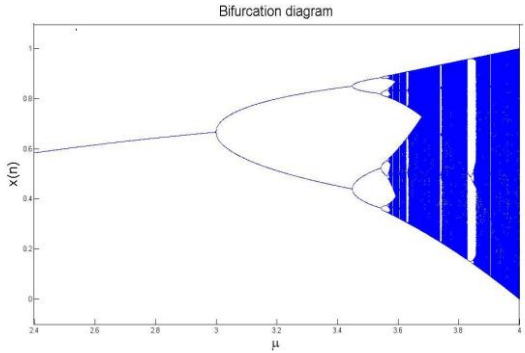

Fig. 2. Bifurcation diagram for logistic map

*2.2. Chua Circuit*

Chua' Circuit is one of the simplest types of chaotic systems (see Fig. 3.a and Fig. 3.b).

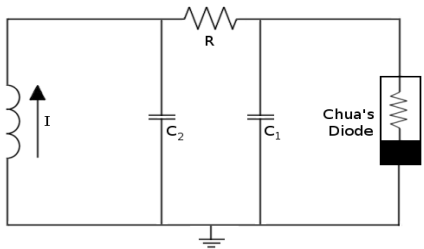

Fig. 3.a. Chua Circuit

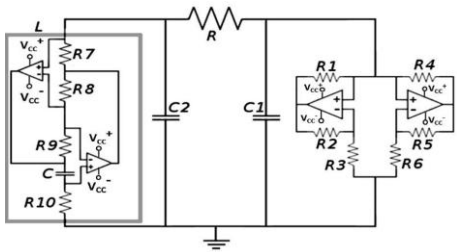

Fig. 3.b. Chua Circuit,
(Notice that the inductor and Chua's diode (nonlinear resistance) are replaced by their equivalent OpAmp circuits

Using an oscilloscope one can watch a Chua circuit creating the chaotic double scroll (Fig. 4), which can be modeled by three equations [11], which are

$$C_1 \, dv_1/dt = (v_2 - v_1) / R - g(v_1) \quad (4)$$

$$C_2 \, dv_2/dt = - (v_2 - v_1) / R + I \quad (5)$$

$$L \, dI/dt = - r \, I - v_2 \quad (6)$$

Where $v_1$ and $v_2$ are voltages across $C_1$ and $C_2$ respectively, g(v1) is the conductance of the nonlinear resistance (the equivalent of Chua's diode) and r is inductor's resistance [11].

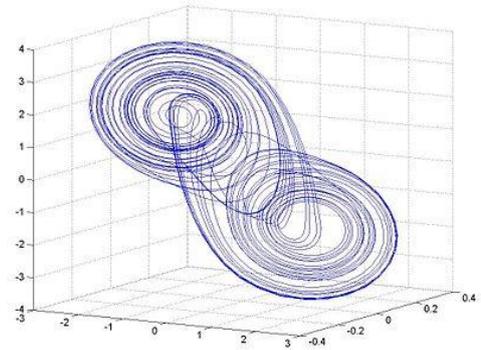

Fig. 4. Double-scroll chaos

Since the goal is to control the chaotic behavior of the circuit above by using Artificial Neural Networks (ANN), the next section will be devoted to discussing various aspects of ANN.

## 3. Artificial Neural Network (ANN)

ANN is part of Artificial Intelligence (AI) which emphasizes the creation of intelligent machines that work and react like humans [12]. It is based on the biological neural system. The Information that flows through the network affects the structure of the ANN because a neural network changes or learns, in a sense based on that input and its output's feedback. Fig. 5 shows an example of an Artificial Neuron and Fig. 6 shows an example of an Artificial Neural Network [12].

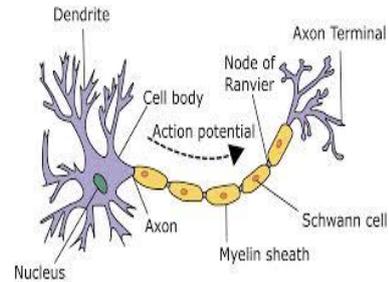

Fig. 5. An example of an Artificial Neuron

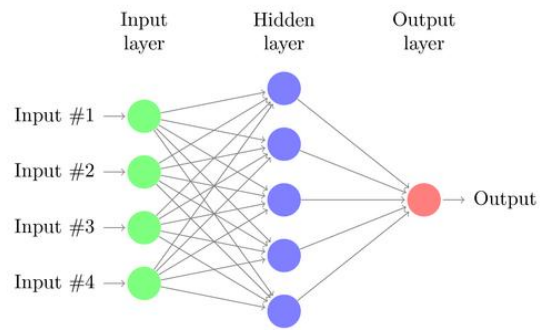

Fig. 6. An Artificial Neural Network (ANN)

An Artificial Neuron could be represented as in Fig. 7, where its output is governed by the following equation:

$y = f(v)$, where f(v) is the activation function and
$v = w_1 \, x_1 + w_1 \, x_1 + \ldots + w_m \, x_m + w_0 \, b_0$

where
$w_0, w_1, w_2, \ldots w_m$ are the weights,
$x_1, x_2, \ldots x_m$ are the inputs, and $b_0$ is the bias

To get some desired output $y_d$, we propagate the neurons' outputs back to the system and that will adjust the weights so as to get the desired output. Fig 7.1.a shows a simple neuron system and fig 7.1.b shows the backprobagation (feedback) system.

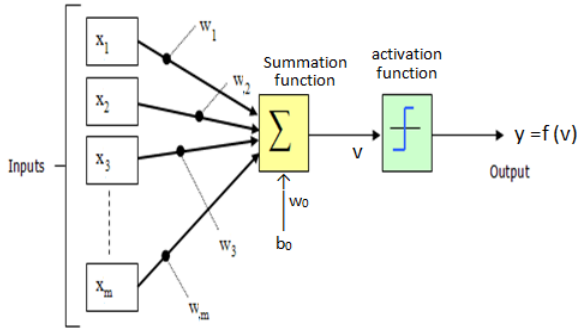

Fig. 7.a. An example of an Artificial Neuron showing its Inputs, weights, summing function and activation function

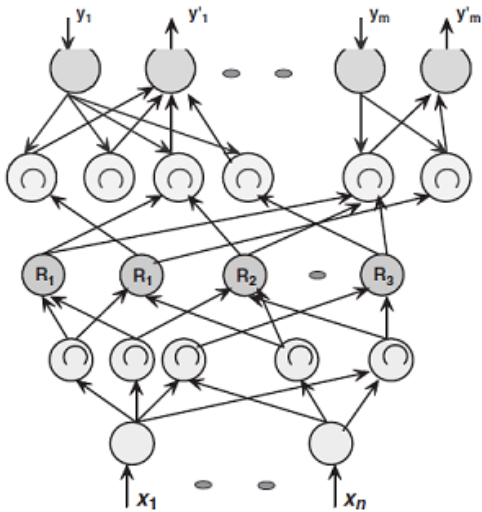

Fig. 7.b. An example of a backprobagation system.

There are many types of activation functions (see [8] for many examples). The activation function we use here is the sigmoid function (Fig. 8).

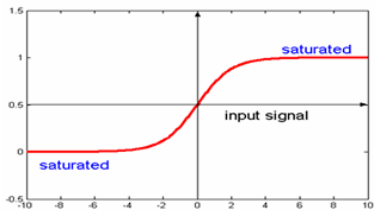

Fig. 8. Activation function; Sigmoid,
$f(x) = 1 / (1 + e^{-x})$, $-\infty < x < \infty$

## 4. Using ANN to Control Chaos

To control the chaotic behavior exhibited by Chua circuit, we first built Chua circuit on a breadboard (Fig. 9), tested it and got the output shown on an oscilloscope (Fig. 10).

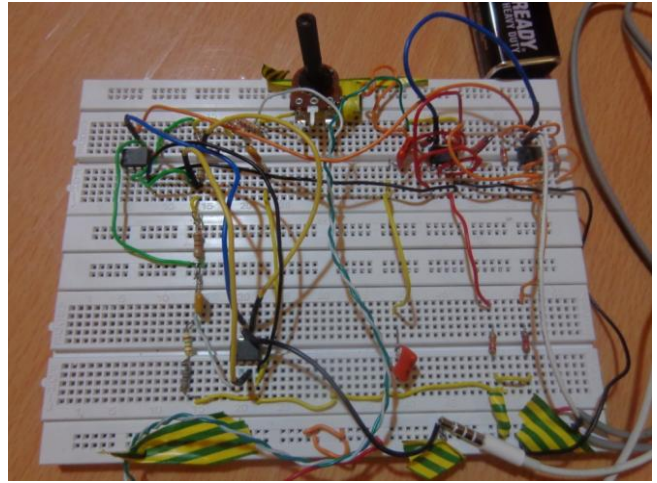

Fig. 9. Chua circuit we built on a breadboard

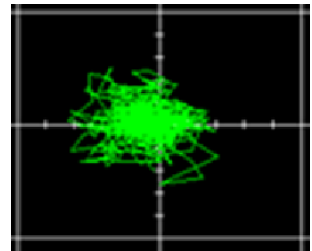

Fig. 10. The chaotic output of the circuit we built

By adjusting the values of the resistance R and C (see Fig. 3.b) it was possible to get rid of the chaotic behavior and forced the circuit to give some desired stable outputs (see Fig. 11.a thru Fig. 11.E for response to different choices of R and C).

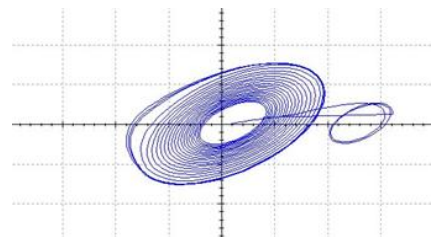

Fig. 11.a. The output response to the first choice of R and C

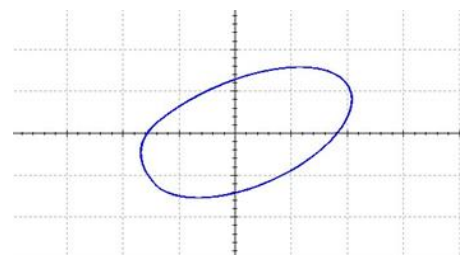

Fig. 11.b. The output response to the second choice of R and C

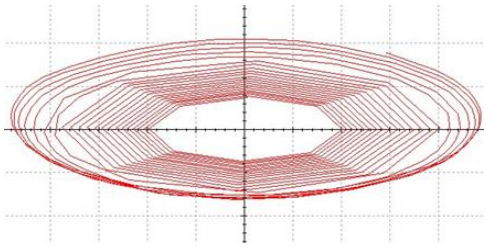

Fig. 11.c. The output response to the third choice of R and C

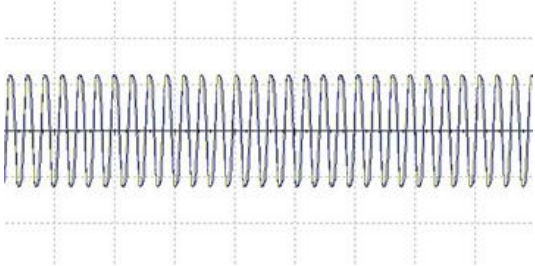

Fig. 11.d. The output response to the fourth choice of R and C

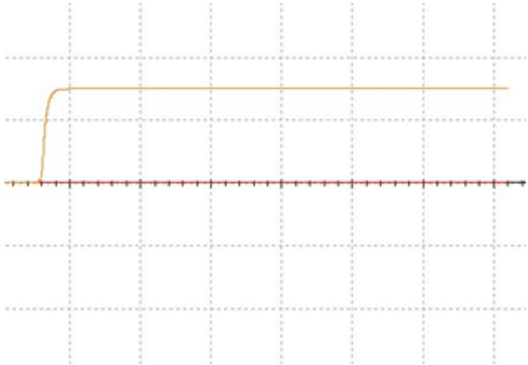

Fig. 11.e. The output response to the fifth choice of R and C

MatLab was also used to simulate the circuit above, using the available Chua circuit's program [6] for its equations, and the expected chaotic behavior happened (Fig. 12.a). By adjusting the system's parameters we ended up with outputs showing no chaotic behavior.

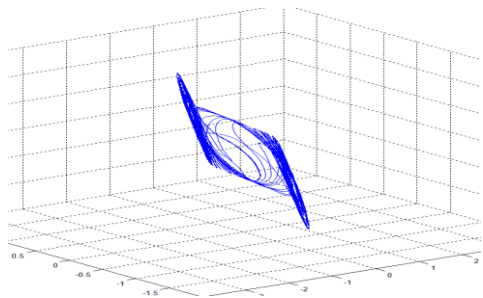

Fig. 12. The output of the MatLab simulation; chaotic behavior

After getting the chaotic behavior by using our electronic circuit built on a breadboard and then by using MatLab available program, VI's Multisim program was used to redraw the electronic circuits for the individual parts of ANN (weights, summing functions, activation functions) (as in [13]) and then all circuits were connected together. Finally, the output of the ANN big circuit was connected to Chua circuit, and part of Chua's circuit output was feedback to the ANN to adjust the weights accordingly.

Once all circuits were connected together (Fig.13), it was possible to use different initial values for the ANN weights and then let the ANN, through learning, adjust Chua circuit's parameters so it gives some desired output instead of the Chaotic behavior which it usually manifests. Samples of the output are shown on Fig. 14.a thru Fig.14.c.

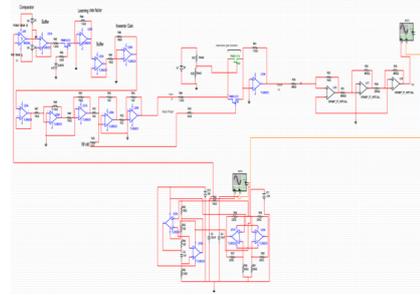

Fig.13. All ANN circuits connected together and their final output connected to Chua circuit

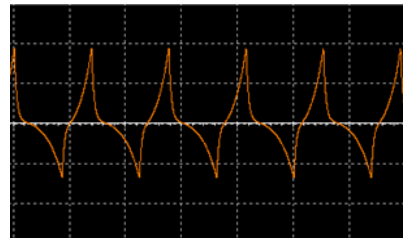

Fig.14.a Sample (a) of the output of the ANN controlled Chua circuit

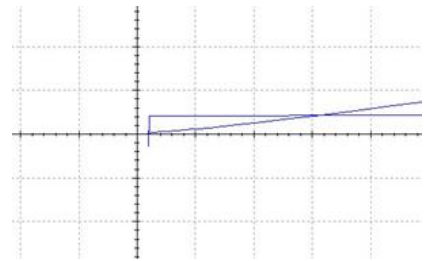

Fig.14.b Sample (b) of the output of the ANN controlled Chua circuit

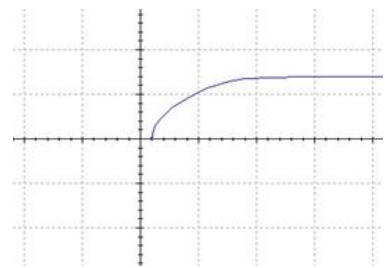

Fig.14.c Sample (c) of the output of the ANN controlled Chua circuit

## 5. Conclusions

Using Artificial Neural Networks (ANN) proved to be very effective in controlling Chaos; that is getting some desired outputs out of chaotic systems. It was also shown that using ANN could lead to stabilizing chaotic systems. The main problem in accomplishing such desired results was the time taken to adjust the ANN weights, first manually one by one then through adjusting Chua's output resistance, then automatically by the feedback control system. This work could be improved by combining ANN control techniques with some other intelligent control techniques such as Fuzzy Logic and Genetic Algorithms, to get the benefit of all.

## 6. References

1. Chen, Heng-Hui, et al. "Chaos synchronization between two different chaotic systems via nonlinear feedback control." *Nonlinear Analysis: Theory, Methods & Applications* 70.12 (2009): 4393-4401.
2. Cai, Guoliang, and Zhenmei Tan. "Chaos synchronization of a new chaotic system via nonlinear control." *Journal of Uncertain Systems* 1.3 (2007): 235-240.
3. Xu, Chang-Jin, and Yu-Sen Wu. "Chaos control and bifurcation behavior for a Sprott E system with distributed delay feedback." *International Journal of Automation and Computing* 12.2 (2015): 182-191.
4. Yoo, Sung Jin, Jin Bae Park, and Yoon Ho Choi. "Stable predictive control of chaotic systems using self-recurrent wavelet neural network." *international journal of control, automation, and systems* 3.1 (2005): 43-55.
5. Liu, Bo, et al. "Improved particle swarm optimization combined with chaos." *Chaos, Solitons & Fractals* 25.5 (2005): 1261-1271.
6. Bishop, Robert C. "What could be worse than the butterfly effect?." *Canadian Journal of Philosophy* 38.4 (2008): 519-547.
7. Heylighen, Francis, Paul Cilliers, and Carlos Gershenson. "Complexity and philosophy." *arXiv preprint cs/0604072* (2006).
8. Sparrow, Colin. *The Lorenz equations: bifurcations, chaos, and strange attractors*. Vol. 41. Springer Science & Business Media, 2012.
9. Liapunov Exponent, http://math.uhcl.edu/shiau/Course/MATH4136(2011)/LiapunovV2 pdf. Accessed 11 Sept. 2016.
10. Li, Damei, et al. "Estimating the bounds for the Lorenz family of chaotic systems." *Chaos, Solitons & Fractals* 23.2 (2005): 529-534.
11. Elwakil, A. S., and M. P. Kennedy. "Chua's circuit decomposition: a systematic design approach for chaotic oscillators." *Journal of the Franklin Institute* 337.2 (2000): 251-265.
12. Rashid, Tariq. *Make Your Own Neural Network*, 2016, Amazon Edition.
13. Khuder, A. I., and Sh H. Husain. "Hardware Realization of Artificial Neural Networks Using Analogue Devices." *Al-Rafadain Engineering Journal* 21.1 (2013).
14. The Stanford Encyclopedia of Philosophy. http://plato.stanford.edu/entries/chaos. Accessed 11 Sept. 2016.